# Digital sovereignty in professional translation workflows: a case study of migrating from Windows to Linux

**Michael Farrell**

IULM University, Milan, Italy

michael.farrell@iulm.it

**Abstract**

This study examines digital sovereignty in professional translation through a screening of computer-assisted translation tools and translation management systems, and a case study of migrating a translator's workstation from Windows to Linux. The framework combines three principles – Control, Continuity and Exit – with three practical requirements: native Linux operation, independence from mandatory external services, and independent access to essential data and resources. Drawing on documentary evidence and the limited number of responses received from providers, the ecosystem screening distinguishes Linux accessibility from operational independence: browser-based platforms may remain dependent on provider-controlled infrastructure. The migration case demonstrates the feasibility of reconstructing one professional workflow around a locally operable Linux core, using LoopCAT, LibreOffice and locally run TranslateGemma, alongside tools for optical character recognition, communication, credential management and backup. Selected cloud services remained available where needed. The findings suggest that digital sovereignty depends on how workflow dependencies are organized, rather than on individual applications or operating systems. It does not require technological isolation, but meaningful control over essential components, continuity of work and practical exit options. The framework is tailored to the case examined here and may require adaptation to other professional workflows.

## 1 Introduction

### 1.1 The Windows 10 problem

Windows 10 was scheduled to reach its end of support on 14 October 2025 (Microsoft, 2025), after Microsoft had announced Windows 11 in 2021 (Microsoft, 2021). For hardware that could not meet Windows 11 requirements, continued use meant accepting an unsupported system, purchasing extended support or replacing otherwise functional hardware. Following consumer advocacy, Microsoft provided consumers in the European Economic Area with a free one-year extension of Extended Security Updates (Euroconsumers, 2025). This prospect nevertheless prompted the author to question how much control professionals truly have over their computing environments.

### 1.2 Digital sovereignty

Digital sovereignty is often discussed at the level of states and political communities, although some academic literature also considers the autonomy of individuals, workers and organizations. In the European context, Madiega (2020) discusses digital sovereignty as a strategic objective linked to Europe's capacity to control its digital environment and strengthen its technological autonomy. Floridi (2020) extends the discussion to control over data, software, standards, hardware and digital

Preprint of paper submitted for formal peer review at the 47th Translating & the Computer conference, Luxembourg, 8–10 December 2026.

infrastructure; Roberts et al. (2021) emphasize the importance of legitimate control over digital resources; and Wang et al. (2026) also consider digital sovereignty at the level of individuals, workers and consumers. These approaches provide a conceptual foundation but offer limited guidance on how to assess whether an individual professional workflow remains controllable when a provider changes its requirements or withdraws support.

Digital Independence Day (DI.DAY), a grassroots initiative coordinated by Save Social – Networks For Democracy, encourages people to reduce their dependence on dominant technology providers by switching to alternatives. Although its focus is broader than professional translation and its approach is primarily practical and advocacy-oriented, it reflects related concerns with user choice, control and dependence on technology providers (Digital Independence Day, n.d.).

### 1.3 Digital sovereignty in professional translation: research gap

Professional translation is a useful test case for digital sovereignty because its workflows depend on an interconnected set of technologies, including operating systems, computer-assisted translation (CAT) tools, translation memories and terminology systems, office software, MT/AI services and online platforms. Dependencies can arise not only from the software itself but also from licensing models and operating-system constraints. For example, CAT tool vendors have increasingly moved towards subscriptions: memoQ offers Translator Pro subscriptions to new customers while existing customers can retain perpetual licences (memoQ, 2024), and RWS no longer offers new perpetual licences for Trados Studio Freelance (RWS, 2026). At the same time, major commercial CAT tool vendors have historically not provided native Linux versions (Linux for Translators, n.d.), limiting the operating systems available to users. Browser-based products such as Trados Go (RWS, 2025) and memoQweb (memoQ, 2026) offer a partial workaround by enabling access from different operating systems, but this flexibility comes at the cost of greater reliance on externally operated services.

Translation-technology research has examined CAT tools and their role in professional translation workflows (Bowker, 2015), alongside efforts to improve translation environments through technical and interface-level developments (Vandeghinste et al., 2019). Related work has explored translator autonomy and translation-data governance (Vieira et al., 2021; Moorkens & Lewis, 2019). However, less attention appears to have been paid to how these concerns can be brought together in a practical assessment of control, continuity and portability within an individual professional workflow.

This study addresses the apparent gap between the broad conceptual literature on digital sovereignty and the practical evaluation of professional translation workflows through a two-stage investigation: (1) a structured *CAT Tool/translation management systems (TMS) Digital Sovereignty Survey*, hosted on EUSurvey, and documentary screening; and (2) a case study of migrating a professional workstation from Windows to Linux.

### 1.4 Research questions and contribution

The study asks four questions:

- RQ1. How can digital sovereignty be conceptualized and assessed in an individual professional translation workflow?
- RQ2. To what extent could the author's professional translation workflow be migrated from Windows to Linux while retaining the functionality required for his professional work?
- RQ3. What technical and professional-practice constraints on digital sovereignty emerge during the migration?

- RQ4. To what extent do the translation technologies examined meet the proposed criteria for digital sovereignty?

The contribution is a practical framework for assessing digital sovereignty in a professional translation workflow, together with a case study that applies it to a Windows-to-Linux migration and identifies concrete technological dependencies

## 1.5 Conceptual framework and evaluation criteria

### 1.5.1 Working definition

For this study, *digital sovereignty is the capacity to exercise effective control over the digital infrastructure, software, services and data on which professional activity depends, while retaining the ability to continue operating and to migrate to alternatives without being subject to unilateral decisions by essential external providers.*

### 1.5.2 Core principles

Three principles give the concept practical meaning. **Control** concerns authority over software, services, data and language resources, rather than independence from every external technology. For example, a locally stored translation memory offers more direct control than a cloud platform from which data cannot readily be exported. However, control may also be limited when an LSP assigns freelancers to a proprietary environment containing shared memories and terminology databases (Smartling, n.d.).

**Continuity** concerns the ability to keep working despite changes in OS support, licensing, product availability, cloud services, file formats or access to externally provided language resources. The end of support for Windows 10, discussed in the introduction, illustrates why continuity matters.

**Exit** is the ability to migrate without unacceptable loss of data, functionality or productivity. It depends on systems working together, open standards, data portability and manageable switching costs. Export alone is insufficient if project files and language resources cannot be used effectively elsewhere. This definition of Exit draws on research linking data portability to compatibility and switching costs (Jeon et al., 2023), as well as work on reversibility and exit-by-design (Wang et al., 2026).

Dependencies are therefore not inherently problematic. An ordinary dependency can be replaced, bypassed or migrated at acceptable cost. A critical dependency is essential, subject to unilateral external change and lacks a practical exit path.

Local-first software supports these principles by prioritizing local data ownership and control without sacrificing the benefits of synchronization and collaboration (Kleppmann et al., 2019).

### 1.5.3 Hard gates

The practical circumstances of the case study shaped the initial selection criteria. The workstation could not be upgraded to Windows 11, making a transition away from Windows necessary, and Linux was selected as the alternative operating system.

The hard gates provide minimum eligibility conditions for preliminary screening, while Control, Continuity and Exit support a broader assessment of digital sovereignty. Optional cloud and AI functions did not automatically disqualify a tool; the key question was whether an external service was indispensable for essential work. The criteria were developed for the specific workflow examined in this study and provide a basis for evaluating other professional contexts. Their application elsewhere may involve adapting the technical criteria to the requirements of each workflow, while retaining the underlying principles.

| Hard gate | Operational requirement |
|---|---|
| Native Linux operation | Essential workflow functionality operates on Linux without Windows or macOS. |
| No mandatory external service | Essential functionality does not require continuing access to an externally controlled service. |
| Independent access to essential data and resources | Essential project data and language resources remain accessible, exportable and usable independently of a provider. |

Table 1: Operational criteria for the digital sovereignty hard gates

## 2 Methods

### 2.1 Research design

The first stage screened CAT/TMS software using the *CAT Tool/TMS Digital Sovereignty Survey* and public documentation. The survey covered platform compatibility, data management and portability, cross-system functionality, licensing, online requirements, offline operation, MT/AI dependencies, vendor access, deletion after termination, data protection, jurisdiction and security. These items were interpreted through Control, Continuity and Exit rather than treated as separate pass/fail criteria. Providers were contacted directly, but few responded; non-response was not treated as failure. Tools were excluded only where the available evidence established failure of a hard gate; otherwise, the relevant status was recorded as not established.

The screening combined provider responses with publicly available documentation to identify potential dependencies and tools warranting further evaluation. The depth of evidence varied between products, and the results should be understood as a structured preliminary assessment rather than a comprehensive technical audit.

The second stage examined the author's migration of his professional workstation from Windows to Linux, comparing the resulting workflow with the pre-migration setup. A backup Windows 11 laptop remained available to maintain continuity during the transition. Where practicable, cross-platform tools were selected to maintain similar working environments on both machines and reduce disruption when switching between them.

To reduce the learning curve, the author chose a Linux environment that would be familiar to Windows users, drawing on migration experiences shared on YouTube and blogs. Two candidate Linux distributions were tested by booting from live USB drives to compare their interfaces and check basic hardware compatibility before installation. Linux Mint with Cinnamon was selected following these preliminary tests. It was initially installed alongside Windows 10 in a dual-boot configuration, allowing further hardware and workflow testing before the migration was completed. After this trial, Linux Mint was reinstalled as the workstation's sole operating system.

An alternative would have been to install Linux on an external solid-state drive, leaving the internal drive unchanged. This approach could be particularly suitable for users less confident with technical changes, as it allows them to try Linux while keeping their existing system intact.

### 2.2 Baseline workflow

The baseline workflow included Déjà Vu X, memoQ and Trados Studio. Microsoft 365, especially Word, was essential for handling DOCX files, while PowerPoint and Excel were used less often. Selectable-text PDFs were opened in Word, whereas scanned PDFs were processed using optical character recognition (OCR) provided by the Canon IJ Scan Utility. Communication relied on Thunderbird, Zoom and WhatsApp, and local backups provided redundancy. The DeepL application programming interface (API) was integrated into the CAT environment, while ChatGPT was accessed through a browser.

The normal sequence is shown in the table below.

| Stage | Workflow activity |
|---|---|
| 1 | Receipt and storage of source files |
| 2 | Document preparation |
| 3 | Import into CAT environment |
| 4 | Translation and revision, with MT/AI where appropriate |
| 5 | Generation of target files |
| 6 | Quality assurance |
| 7 | Delivery to client |
| 8 | Backup of project files |

Table 2: Stages of the baseline translation workflow

Functional testing covered CAT project creation, translation, revision and export; Office and HTML compatibility; MT/AI; PDF/OCR; communication; and basic backup checks. The sovereignty assessment applied the hard gates and the principles of Control, Continuity and Exit. A dependency was considered potentially critical if it was essential, externally changeable and lacked a practical exit.

Security was not systematically evaluated: routine checks cannot establish antivirus effectiveness, and backup reliability can only be assessed fully by testing recovery under failure conditions.

| Category | Windows baseline |
|---|---|
| CAT | Déjà Vu X; memoQ; Trados Studio |
| MT/AI | DeepL API; ChatGPT |
| Office | Microsoft 365, especially Word; PowerPoint and Excel less frequently |
| Web | Firefox browser for dictionaries, terminology databases, corpora and other language resources |
| Communication | Thunderbird; Zoom; WhatsApp |
| PDF/OCR | Microsoft Word for text-based PDFs; Canon IJ Scan Utility for scanned PDFs |
| Backup | Local backup |
| Security | Password manager; Microsoft Defender Antivirus and Firewall |

Table 3: Software environment before migration

### 2.3 Migration and evaluation

The evaluation used representative workflow tasks and separated functional testing from sovereignty assessment.

The migration demonstrated that selected Linux tools could perform the essential tasks required by this professional workflow. The evaluation focused on functional feasibility rather than systematic comparisons of productivity, user satisfaction or performance on complex documents and large projects. These aspects, together with the suitability of the tools for other professional contexts, will become clearer through ongoing use and further evaluation.

| Area | Question |
|---|---|
| CAT workflow | Can projects be created/opened, translated, revised and exported? |
| File compatibility | Can Word, PowerPoint, Excel and HTML be processed without unacceptable structural/formatting loss? |
| MT/AI | Can required functions be performed with available services or local alternatives? |
| PDF/OCR | Can scanned and other PDFs be converted to usable Word files? |
| Communication | Can communication with clients/colleagues continue? |
| Backup | Can working files be backed up reliably? |
| Security | Can relevant security functions be maintained? |

Table 4: Functional evaluation tasks and questions

## 3 Results

### 3.1 CAT/TMS ecosystem

Only three of the twenty-one software providers contacted returned completed questionnaires: Hook & Hatton (Local AI Translator), Wordscope and Maxprograms (Swordfish). According to Hook & Hatton, Local AI Translator is a desktop application for Linux and Windows that can operate offline, supports export in standard formats and uses a one-time licence. The provider reported that vendor personnel cannot access project data and that no user data is stored. Maxprograms reported that Swordfish 5.20.0 is an open-source desktop application for Linux, macOS and Windows that remains fully functional offline after installation and activation. It supports full export of project data in standard formats, and users can choose their AI/MT providers or disable external integrations. On the basis of these responses, both tools met the hard gates. Wordscope, by contrast, was reported to require a continuous internet connection for core functionality. Although the provider confirmed full data export and deletion after contract termination, the tool failed the mandatory-external-service gate.

The documentary screening identified three groups. Windows-dependent desktop products failed the native-Linux gate, while browser-based and cloud systems failed the external-service gate where core functionality depended on provider infrastructure. Browser access from Linux did not, by itself, establish operational independence (memoQ, 2026; Phrase, n.d.). Although Transifex advertises on-premises deployment (Transifex, n.d.), the available documentation did not establish that a locally deployable CAT solution was generally available to individual translators. Other hosted platforms were assessed on the same basis.

The locally installed candidates were retained for further consideration. Screening examined Linux compatibility, Office-format support, local operation and data control, reliance on external services, licensing and maintenance. Gtranslator was excluded as a general-purpose CAT tool candidate because it is designed primarily for editing gettext Portable Object files.

Licensing, activation and registration requirements were assessed under the principle of Continuity, distinguishing between requirements for initial activation and those requiring continued online access to maintain functionality.

The retained pool was not a ranking of products, nor did every candidate undergo equivalent practical testing. It identified tools potentially suitable for further investigation against the requirements of this particular workflow. The subsequent selection and migration tests focused on whether a candidate could perform essential translation tasks in practice while meeting the study's sovereignty criteria.

| Tool/platform | Linux | Mandatory external service | Data continuity/control | Screening |
|---|---|---|---|---|
| Trados Studio | No | – | – | Excluded |
| memoQ translator pro | No | – | – | Excluded |
| Déjà Vu X | No | – | – | Excluded |
| Across Translator Edition | No | – | – | Excluded |
| Phrase | Accessible | Yes | – | Excluded |
| XTM Cloud | Accessible | Yes | – | Excluded |
| Smartling | Accessible | Yes | – | Excluded |
| Lokalise | Accessible | Yes | – | Excluded |
| Crowdin | Accessible | Yes | – | Excluded |
| Smartcat | Accessible | Yes | – | Excluded |
| Matecat | Accessible | Yes | – | Excluded |
| Wordbee Translator | Accessible | Yes | – | Excluded |
| Transifex | Hosted accessible; on-premises advertised | Hosted: yes | Not established for individual local use | Hosted solution excluded |
| OmegaT | Yes | No | Yes | Retained |
| Wordfast Pro | Yes | No | Yes | Retained |
| CafeTran Espresso | Yes | No | Yes | Retained |
| BasicCAT | Yes | No | Yes | Retained |
| Fluency Now 2024 | Yes | No evidence | Not fully established | Retained |
| Heartsome Translation Studio 8.0 | Yes | No | Yes | Retained |
| Supervertaler Workbench | Yes | No | Yes | Retained |
| LoopCAT | Yes; local-first | No | Yes | Retained |
| Local AI Translator | Yes | No | Yes | Retained |
| Swordfish | Yes | No | Yes | Retained |

Table 5: CAT tool screening against the digital sovereignty hard gates

### 3.2 Migration results

#### 3.2.1 Selection of the principal CAT environment

BasicCAT was not prioritized because the available documentation did not confirm direct handling of all three Office formats required by the workflow (BasicCAT, n.d.). Heartsome Translation Studio 8.0 was excluded from practical testing because its discontinued development raised concerns about maintenance and continuity (L10nSoftware, 2016). Supervertaler Workbench was not prioritized because, at the time of screening, its developer stated that it was no longer actively developed (Supervertaler, n.d.). Subsequent repository activity shows that the project has continued to receive updates, although this does not in itself establish that active feature development has resumed.

Wordfast Pro was included in the screening because the vendor lists Linux as a supported platform. However, it could not be made to run on the study system despite following the vendor's installation and troubleshooting instructions, so its Linux functionality could not be evaluated (Wordfast, n.d.).

Local AI Translator was selected for further assessment because it supports local, offline translation through an interactive, segment-based interface (Hook & Hatton, n.d.). It can use existing translation memories in translation memory eXchange format to provide additional context through retrieval-augmented generation (Lewis, 2025). However, since its Linux version is available only on request rather than as a general download (Hook & Hatton, n.d.), Local AI Translator was not prioritized for practical testing.

Five tools remained plausible candidates for further assessment. LoopCAT was selected for practical testing because its local-first architecture, local AI support and Apache-2.0 licence aligned with the study's principles. OmegaT can also use locally hosted LLMs through third-party plugins (Nic-J, n.d.; paashkaa, n.d.), introducing dependencies whose maintenance and compatibility depend partly on the plugin developers. Fluency Now 2024 supports MT engines and API services, but the available documentation did not establish native integration with local LLMs (Western Standard, n.d.). Swordfish supports offline operation after installation and activation, and its open-source code provides an alternative to relying exclusively on the vendor's distributed builds. However, its licensing arrangements may still create dependencies, while building or maintaining the software independently could require additional technical expertise. Local LLM integration through Ollama was introduced in a later version than the one initially screened (Maxprograms, n.d.). LoopCAT's permissive licence does not guarantee continued maintenance, and its maturity remains a limitation. Its selection reflected its relevance to the research question, not superiority in every respect or any failure of the other candidates.

Dogru and Martín Mor (2026) describe LoopCAT as an ongoing open-source project with areas requiring further work, including complex-document fidelity, performance on large projects and release testing across operating systems. As this study tested only simple projects, it does not establish LoopCAT's reliability with complex documents or larger workloads. Its ongoing development and these documented limitations therefore warrant cautious further evaluation.

### 3.2.2 Office and document processing

LibreOffice became the principal office suite. It is used to open and save DOCX, XLSX and PPTX files, preserving client-facing exchange formats without requiring Microsoft 365. The LibreOffice documentation acknowledges that complex layouts and formatting may not be reproduced identically when Microsoft Office documents are converted, while recent releases continue to improve compatibility with these formats (The Document Foundation, 2026; Vignoli, 2026). Compatibility is therefore treated as a functional issue to be checked in practice, rather than assumed from format support alone.

GImageReader replaces Canon IJ Scan Utility, which is not available for Linux.

### 3.2.3 Machine translation and LLMs

DeepL and ChatGPT remain available through Firefox as optional external services. TranslateGemma runs locally through both Ollama and LM Studio. Ollama was initially installed to use TranslateGemma through Supervertaler Workbench. Although the documentation indicates compatibility, the author was unable to get Ollama to work with LoopCAT and therefore added LM Studio as an alternative. Both remain installed, although the author prefers Ollama for standalone use.

#### 3.2.4 Communication, collaboration and web access

Zoom is accessed through its browser interface, while WhatsApp is used on the author's smartphone because clients and contacts use it. Thunderbird was adopted as an email client and runs natively on Linux. Firefox was chosen for web browsing.

#### 3.2.5 Security

Credential management was retained through KeePass Password Safe, which stores an encrypted local database and had already been used by the author in the Windows environment. KopiaUI is used for backups and provides deduplication, compression and authenticated encryption. It supports backups to local storage and to remote locations accessed through Secure Shell (SSH), a protocol for secure communication between computers over a network.

Regarding antivirus protection, ClamUI, a graphical interface for ClamAV, was installed, and regular malware scans are planned. Linux Mint's firewall was also checked and found to be active.

These measures support protection of personal data under the GDPR's risk-based requirements for confidentiality, availability and recovery, but do not by themselves establish compliance. The GDPR prescribes no specific products; protection also depends on configuration, maintenance, tested recovery and the handling of personal data shared with external services (European Parliament and Council, 2016, Article 32).

The resulting architecture had three dependency classes: (1) locally controlled core applications; (2) optional external services, with local AI as an alternative; and (3) externally determined communication services such as Zoom and WhatsApp. The core workflow operated locally without Windows, while external services remained where useful or professionally necessary.

#### 3.2.6 Resulting software environment

| Function | Migrated implementation | Dependency interpretation |
|---|---|---|
| CAT/projects | LoopCAT, local-first | Core/local; basic DOCX, XLSX, PPTX import/export tested successfully |
| Office | LibreOffice | Local; retained DOCX/XLSX/PPTX handling |
| PDF/OCR | GImageReader; Word-compatible editable output | Local OCR; PDF→OCR/editable document→CAT |
| MT/AI | DeepL, ChatGPT; TranslateGemma via Ollama/LM Studio | External services optional; local alternative available |
| Email | Thunderbird | Native Linux/local |
| Meetings | Zoom via browser | External communication service |
| Messaging | WhatsApp via smartphone | Externally determined communication platform |
| Browser | Firefox | Linux/open source |
| Credentials | KeePass Password Safe | Encrypted local database; no mandatory cloud |
| Backup | KopiaUI | Local/SSH, encrypted; no mandatory cloud |
| Antivirus | ClamAV via ClamUI | Local malware scanning; detection depends partly on the freshness of externally updated signatures |
| Firewall | Linux Mint firewall | System firewall checked and found active; no mandatory commercial security service |

Table 6: Software environment after migration to Linux

## 4 Discussion

The case study suggests that digital sovereignty depends on the architecture of workflow dependencies rather than on individual applications. Linux compatibility alone is not sufficient if essential data or functions remain under external control. Conversely, external services can be compatible with digital sovereignty when their use is optional and essential work does not depend on them. A local-first approach therefore does not seek to eliminate external services, but to ensure that essential work can continue without relying entirely on them.

Sovereignty is also constrained by hardware and technology infrastructure. Devices may depend on proprietary drivers or closed-source firmware even when much of the software is open source; the Linux kernel project notes that some firmware is distributed only in binary form (Linux Kernel Project, 2024).

In this case, **Control** was strengthened by moving the core CAT environment, project data and language resources to local Linux software, although some external dependencies remained. **Continuity** was maintained by preserving essential workflow functions, including Office-compatible file handling, CAT-tool functionality, email, web access, AI/MT and backup. **Exit** was demonstrated by transferring the workflow to an alternative operating system, although its long-term sustainability depends on continued cross-system compatibility and the availability of portable file formats. This practical interpretation is consistent with the practitioner-oriented LibreInfra discussion "Digital sovereignty is the ability to change course" (2026), which emphasises the importance of retaining the ability to change direction rather than eliminating all dependencies.

The migration also highlights the distinction between nominal and functional portability. DOCX, XLSX and PPTX files formed part of the professional interface with clients, making compatibility important to the continuity of work. However, the ability to export data does not guarantee that it can be used in another system without unacceptable loss. Similarly, accessing a CAT/TMS through a browser does not guarantee independence if core functions remain dependent on provider-controlled infrastructure. In practice, Exit depends on whether essential resources and capabilities can be transferred to an alternative environment without unacceptable disruption.

The case also shows that digital sovereignty is shaped by professional constraints. The backup Windows 11 laptop provided continuity during the migration, while cross-platform tools preserved flexibility and increased control over the working environment on both machines. Some benefits of the migration therefore extended to the backup computer even though it continued to run Windows. At the same time, the practical demands of collaboration limited the author's choice of communication technologies: Zoom and WhatsApp remained externally determined dependencies because clients and contacts relied on them, although neither was critical to the core CAT and data environment.

For translation-technology research, dependency architecture can complement established concerns such as productivity, usability and performance. The hard gates provide a preliminary means of identifying critical dependencies, while Control, Continuity and Exit support a broader assessment of the workflow. Local or open-source tools are not inherently preferable: deployment, data location, export options, licensing, mandatory external services, the ability of systems to work together, and switching costs also shape the degree of control available to the user.

The LoopCAT case also illustrates that local operation and open-source licensing do not, by themselves, guarantee long-term continuity. Development activity, maintenance and the ability to address limitations affect a tool's viability for professional use, particularly in the case of emerging tools whose long-term reliability remains uncertain.

The study has four main limitations: it is a single case involving one translator and workstation; the ecosystem analysis relied partly on provider documentation and only three survey responses; the hard gates are specific to this Windows-to-Linux context; and the study did not systematically compare productivity, usability or user satisfaction across CAT tools.

## 5 Conclusion

This study examined how digital sovereignty can be assessed and put into practice within a professional translation workflow. It applied the principles of Control, Continuity and Exit through a framework that also included three hard gates: native Linux operation, independence from mandatory external services, and independent access to essential data and resources. The framework provides a practical way to examine dependencies at workflow level, complementing established concerns in translation-technology research such as productivity, usability and performance. Its criteria are grounded in the practical requirements of this Windows-to-Linux migration and may provide a basis for assessing digital sovereignty in other professional contexts.

The ecosystem screening showed that Linux accessibility does not necessarily imply operational independence, while local operation alone does not guarantee portability or long-term maintainability.

The migration case demonstrated the practical feasibility of reconstructing the author's professional workflow around a locally operable Linux core, with local alternatives for essential translation and document-processing tasks. Selected external services remained available where useful or necessary for professional communication and collaboration. The case also showed that sovereignty can be improved incrementally: the objective need not be to eliminate every external dependency, but to preserve control over essential work and realistic alternatives when technologies or service conditions change.

Further research could apply and refine the framework across different translators, CAT/TMS environments and professional settings, and evaluate cross-system compatibility, performance, security and maintenance over time. This would help establish the extent to which the practical benefits observed in this case can be generalized.

## Acknowledgements

The research project reported in this paper has received funding from the International Center for Research on Collaborative Translation at IULM University, Milan, Italy.